\documentclass{article} 
\usepackage{iclr2021_conference,times}
\usepackage{makecell}
\usepackage{amsmath,amsfonts,bm}

\usepackage{url}
\usepackage{graphics}
\usepackage{natbib}

\usepackage{graphicx}
\usepackage{subfigure}

\title{Data-driven Weight Initialization with \\ Sylvester Solvers}

\author{Debasmit Das, Yash Bhalgat \& Fatih Porikli \\
Qualcomm AI Research \thanks{Qualcomm AI Research is an initiative of Qualcomm Technologies, Inc.} \\
\texttt{\{debadas,ybhalgat,fporikli\}@qti.qualcomm.com}
}

\iclrfinalcopy 
\begin{document}

\maketitle

\begin{abstract}
In this work, we propose a data-driven scheme to initialize 
{the parameters of a deep neural network.}
This is in contrast to traditional approaches which randomly initialize parameters by sampling from transformed standard distributions. Such methods do not use the training data to produce a more informed initialization. Our method uses a sequential layer-wise approach where each layer is initialized using its {input {activations}}. The initialization is cast as an optimization problem where we minimize a combination of encoding and decoding losses of the {input activations}, which is further constrained by a user-defined latent code. The optimization problem is then restructured into the well-known Sylvester equation, which has fast and efficient \textit{gradient-free} solutions. 
{Our data-driven method achieves a boost in performance compared to random initialization methods, both before start of training and after training is over. We show that our proposed method is especially effective in few-shot and fine-tuning settings. We conclude this paper with analyses on time complexity and the effect of different latent codes on the recognition performance.}
\end{abstract}

\section{Introduction}
Deep neural networks have produced state-of-the-art recognition performance in areas like computer vision~\citep{szegedy2015going,he2016deep}, natural language processing~\citep{devlin2018bert,brown2020language}, speech recognition~\citep{nassif2019speech}, etc. The success of these deep neural network models has been mostly attributed to the quality and quantity of datasets, complex architectures and algorithms and advanced computing resources. However, lesser credit has been attributed to developing novel and effective initialization schemes for these deep and complex architectures. 


The goal of an initializer is to obtain parameters that set the initial state of a neural network into the basin of a good local minimum{~\citep{li2018visualizing}}. 
Since the optimization landscape might contain large number of local minima \citep{auer1996exponentially}, finding the right one becomes difficult and so researchers have chosen random initializers. For example, \citet{krizhevsky2012imagenet} 
initialized the AlexNet weights from a Gaussian distribution with zero mean and 0.01 standard deviation. However, using such initialization for deeper networks causes the gradients or activations to explode or vanish in the extreme layers. 
To take care of exploding and vanishing signals, 
\citet{Glorot} introduced a multiplying factor on the standard deviation of the Gaussian distribution from which the weights are sampled, which depended on the fan-in and fan-out of each layer.
\citet{He} extended this idea for ReLU-based activations. 
Both these methods are currently the standard approach for initializing deep neural network layers.

Alternative methods include orthonormal matrix initialization~\citep{saxe}, which empirically performed better than sampling weights from a Gaussian distribution. ~\citet{mishkin2015all} extended this work by scaling the weights using variance of the batch activations. This is similar to batch normalization~\citep{batch} except the weight initialization part. \citet{walk} also proposed a Random Walk-based initialization scheme where scaling was done such that logarithm of norms of the backpropagated errors were preserved. It is important to note that for these methods, only the normalization is data-dependent and not the weight parameters.

All the above-mentioned methods use random initialization for the weights. We conjecture that initializing the weights using training data might produce better performance. 
\cite{magic} proposed a data-driven initialization scheme using principal components or clustering to initialize the weights. However, it uses additional normalization steps with gradient computation. 
Recently, MetaInit~\citep{metainit} is used to boost an existing initializer by learning the parameter norms such that the scaled parameters lie in locally linear regions with minimal curvature. GradInit~\citep{gradinit} extends MetaInit by using actual training samples instead of randomly sampled data and also provides an upper bound on the gradients to prevent trivial solutions. However, for both these methods, the scales for modifying the norms of the parameters are obtained by minimizing a loss term using gradient descent.

To address these challenges, we propose an \emph{efficient gradient-free} approach to \emph{data-driven} weight initialization. Our approach uses a subset of the training data to feed the input activations of each layer of a neural network. We use these input activations to frame an optimization problem where the weights are optimized to encode and decode the activations properly. This setup is inspired from auto-encoder based greedy pre-training~\citep{hinton,yoshua} to obtain good initial neural network states. However, this method required full-fledged training on the full dataset using gradient descent. In our method, we ignore the non-linear activations and provide \emph{flexibility} to choose the latent code. Infact, the optimal solution is reformulated as the solution of a Sylvester equation which has well-defined solvers~\citep{bartels1972solution} without using gradient descent. {The Sylvester equation has also been used to learn the mapping between features and semantic information for zero-shot learning~\citep{kodirov2017semantic}. But it has not been previously used an intermediate step to initialize each layer of a neural network.} Preliminary results on the image classification task with CIFAR-10 and CIFAR-100 datasets justify the advantage of using Sylvester-equation-based-initialization of deep neural networks.
\section{Proposed Approach}
For data-driven initialization, we have access to labeled training data $\mathcal{D} = \{(\mathbf{x}_i, y_i) \}_{i=1}^N$ as well as the network architecture we plan to initialize. For the sake of efficiency, we use a very small subset of the training data $\widetilde{\mathcal{D}} \subset \mathcal{D}$ to initialize the network. We also assume that we have access to the trainable convolutional and feedforward layers as well as the intermediate activations. This allows us to train each layer in a sequential manner where the input activations are used for initialization and the propagated output activations are used to initialize the next layer.

Before initializing a convolutional layer, we also restructure and reshape the input activations and weights. This is done to convert the convolutional layer into a fully connected one, which will eventually allow us to exploit existing dimensionality reduction techniques elegantly. Let the input to the convolutional layer be $\mathbf{X} \in \mathbb{R}^{h \times w \times c_i \times n}$, where $h$ and $w$ are the height and width of the activation map, $c_i$ is the number of input channels and $n$ is the number of samples used for the initialization. Let a convolution filter height, width and depth be $f_h$, $f_w$ and $c_i$. For an input activation map from a single sample, we would obtain a number of $f_h \times f_w \times c_i$ sized patches over which the filter convolves. This can be repeated over $n$ samples. If $n_p$ be the total number of $f_h \times f_w \times c_i$ sized patches, then $n_p$ will depend both on $n$ as well as stride and padding of the convolutional filter. These patches are then flattened to obtain a reshaped input activation $\mathbf{X}' \in \mathbb{R}^{f_h f_w c_i \times n_p}$. 

Let a convolutional weight be represented as a 4D tensor $\mathbf{W} \in \mathbb{R}^{c_o \times c_i \times f_h \times f_w}$, where $c_o$ is the number of output channels. To enable compatibility with $\mathbf{X}'$, $\mathbf{W}$ needs to be reshaped as $\mathbf{W}' \in \mathbb{R}^{c_o \times f_h f_w c_i}$. Thus, the convolutional layer weights and input activations are restructured to that of a fully connected layer. For a fully-connected layer, the input activations  and weights would be represented as $\mathbf{X} \in \mathbb{R}^{d_i \times n}$ and $\mathbf{W} \in \mathbb{R}^{d_o \times d_i}$, where $d_i$ is the input dimension and $d_o$ is the output dimension. Equivalency of dimensions between convolutional and fully connected layers would be as follows: $f_h f_w c_i \equiv d_i$, $n_p \equiv n$ and $c_o \equiv d_o$.

To produce a good initial weight $\mathbf{W}$, it should be able to encode the input activations $\mathbf{X}$ to an informative latent code  $\mathbf{S} \in \mathbb{R}^{d_o \times n}$, which can then decode the original input activations. For simplicity and for weight-sharing, we set the decoder to be the transpose of the encoder. The choice of $\mathbf{S}$ is flexible and possible options include principal components, Fisher discriminant, one-hot codes, etc. To optimize for $\mathbf{W}$, we want to minimize a combination of encoding and decoding loss as shown below by the convex optimization problem
\begin{equation}
\label{eq:encdec}
\underset{\mathbf{W}}{\text{min}} \ \underbrace{||\mathbf{X}-\mathbf{W}^T\mathbf{S}||_F^2}_{\text{Decoding Loss}} + \lambda \underbrace{||\mathbf{W}\mathbf{X}-\mathbf{S}||_F^2}_{\text{Encoding Loss}}
\end{equation}
where the scalar $\lambda$ weighs the encoding loss. To obtain the optimal $\mathbf{W}$, we take derivative of Eq.~\ref{eq:encdec} with respect to $\mathbf{W}$, set it to $\mathbf{0}$ and re-arrange to obtain the following equation
\begin{equation}
\label{eq:sylv}
\underbrace{\mathbf{S}\mathbf{S}^T}_{\mathbf{A}}\mathbf{W} + \mathbf{W}\underbrace{\lambda \mathbf{X}\mathbf{X}^T}_{\mathbf{B}} = \underbrace{(1+ \lambda)\mathbf{S}\mathbf{X}^{T}}_{\mathbf{C}}.
\end{equation}
Equation~\ref{eq:sylv} is also known as the Sylvester equation when we set $\mathbf{A} = \mathbf{S}\mathbf{S}^T$, $\mathbf{B}=\lambda \mathbf{X}\mathbf{X}^T$ and $\mathbf{C}=(1+ \lambda)\mathbf{S}\mathbf{X}^{T}$. The Sylvester equation can be efficiently solved by the Bartels-Stewart algorithm~\citep{bartels1972solution}, which has a worst-case time complexity of $\mathcal{O}(d_i^3)$ with the assumption that $d_i > d_o$. This suggests that the time complexity of the Sylvester solver is independent of the number of training samples $n$.  However, time-complexity of obtaining the user-defined latent code $\mathbf{S}$ can depend on $n$. {Consequently, the initialization time might be constrained by the amount of training samples used for computing the latent code. This is especially valid for initializing convolution layers, where large number of patches are used as input activations. Since all the patches are not informative of the object present in the image and to improve computational efficiency, it makes sense to select a subset with a fixed number of random patches in an image as the input activations.} Using the activations in the Sylvester solver, the optimal solution $\mathbf{W}^{*}$ is obtained. It is then reshaped into the appropriate dimension if it represents a convolutional layer else it is kept intact. Then, the input activations are processed through this layer to obtain output activations which after passing through non-linearities act as input activations for the next layer. The process is then repeated for the next layer until the final layer initialization is complete.
\section{Experimental Results}
\subsection{Implementation Details}
To evaluate our method, we use the ResNet-20 backbone on the CIFAR-10 and CIFAR-100 datasets. For the optimizer, we use SGD with initial learning rate of 0.1 and momentum of 0.9. Furthermore, the learning rate is decayed by a factor of 10 at epochs 100 and 150 for the CIFAR-10 dataset and at 80 and 120 for the CIFAR-100 dataset. The training is carried out for over 200 epochs.
For our data-driven initialization, we use a subset of the training data, i.e. 100 samples per class for the CIFAR-10 and 10 samples per class for CIFAR-100 dataset unless limited by the few-shot setting. Unless explicitly mentioned, we use $\lambda=10$ to weigh the encoding loss in Eq.~\ref{eq:encdec}. For the target code $\mathbf{S}$, we use principal components for all layers except for the last layer. For the last layer, $\mathbf{S}$ is set as one-hot code. 
\subsection{Comparison Studies}
\begin{figure}[h]
    \vspace{-20pt}
    \centering
    \subfigure[]{\includegraphics[width=0.4\textwidth]{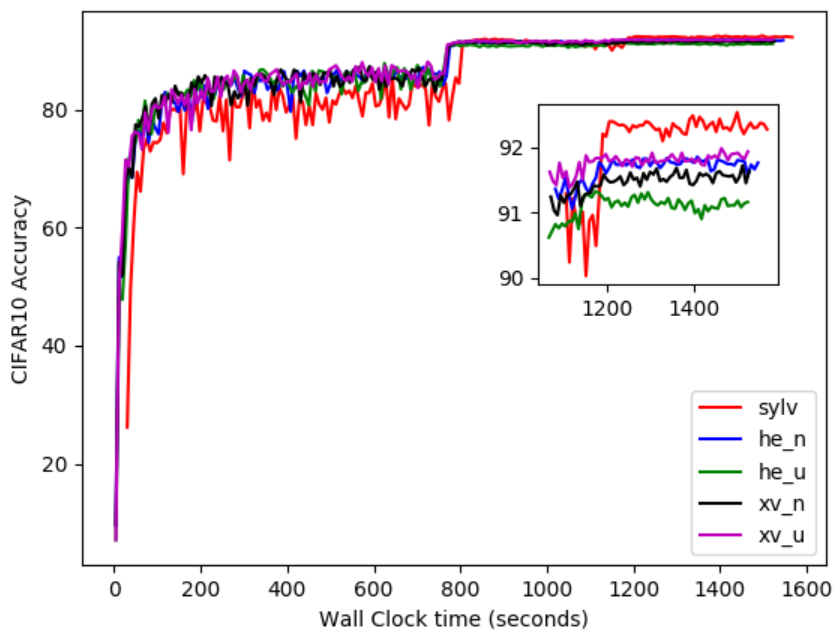}} 
    \subfigure[]{\includegraphics[width=0.4\textwidth]{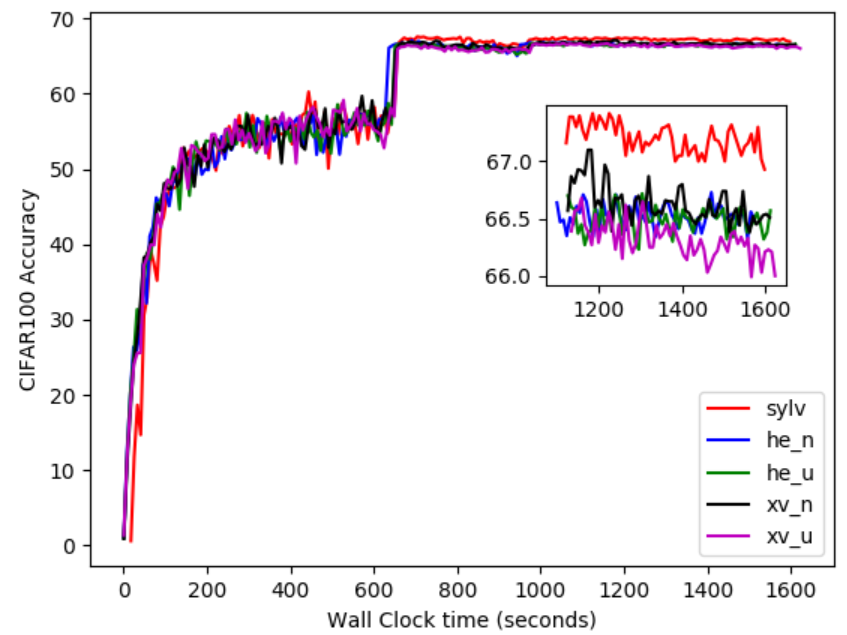}} \vspace{-10pt}
    \caption{Accuracy versus wall clock time comparison with random initialization methods on (a) CIFAR-10 and (b) CIFAR-100 datasets. }\label{fig:initial}
\vspace{-15pt}
\end{figure}

In this subsection, we study the performance of our method compared to random initialization approaches like He uniform, He normal, Xavier uniform and Xavier normal on CIFAR-10 and CIFAR-100 datasets. As shown in Fig.~\ref{fig:initial}(a) and (b), our method produces better final test accuracy compared to random initialization approaches. In terms of initial accuracy, our method produces much better recognition performance (around 25 \%) on the CIFAR-10 dataset. However, for the CIFAR-100 dataset, the initial performance is lower mainly because of larger amount of categories and the use of lesser amount of samples (10) per class for the initialization.
\subsection{Few-Shot Setting}
In this sub-section, we study the effect of different initialization methods in the setting where training samples are few. Specifically, we consider two setups: (a) when the model is trained from scratch after initialization of the whole network, and (b) when a pre-trained model is fine-tuned on a new dataset after only the final classification layer is initialized. For setup (a) we test on both CIFAR-10 and CIFAR-100 using the ResNet-20 architecture while training on few samples. For setup (b) we pre-train the ResNet-20 model on CIFAR-100 and fine-tune on few training samples of CIFAR-10. The number of few-shot samples per class that we consider are 10, 50, 100, 500. The results are shown in Table~\ref{tab:fewshot}.
\begin{table}[h!]
\centering
\caption{Final test accuracy on different few-shot settings with different initialization methods. Numbers in parantheses indicate initial accuracies before start of training. }\label{tab:fewshot}
\vspace{0pt}
\resizebox{\columnwidth}{!}{
\begin{tabular}{|c|c|c|c|c|c|c|c|c|c|c|c|c|}
\hline 
\textbf{Setup} &  \multicolumn{4}{c|}{\textbf{CIFAR-10}}  &  \multicolumn{4}{c|}{\textbf{CIFAR-100}} &  \multicolumn{4}{c|}{\textbf{CIFAR-100} $\rightarrow$ \textbf{CIFAR-10}}\\
\hline
\makecell{Shot $\rightarrow$\\Method $\downarrow$} &  10 & 50 & 100 & 500 & 10 & 50 & 100 & 500  & 10 & 50 & 100 & 500 \\
\hline
\makecell{He\\Uniform} & \makecell{\textbf{27.91}\\(9.96)} & \makecell{40.84\\(9.96)} & \makecell{51.98\\(9.96)} & \makecell{75.20\\(9.96)} & \makecell{11.61\\(1.01)} & \makecell{30.82\\(1.01)} & \makecell{42.52\\(1.01)} & \makecell{67.00\\(1.01)} & \makecell{55.66\\(15.57)} & \makecell{70.64\\(15.57)} & \makecell{74.59\\(15.57)} & \makecell{83.18\\(15.57)}\\
\hline
\makecell{He\\Normal} & \makecell{26.97\\(\textbf{10.0})} & \makecell{43.70\\(10.0)} & \makecell{51.56\\(10.0)} & \makecell{77.31\\(10.0)} & \makecell{10.68\\(0.93)} & \makecell{32.42\\(0.93)} & \makecell{44.51\\(0.93)} & \makecell{66.97\\(0.93)} & \makecell{54.79\\(5.07)} & \makecell{70.82\\(5.07)} & \makecell{75.02\\(5.07)} & \makecell{83.04\\(5.07)} \\
\hline
\makecell{Xavier\\Uniform} & \makecell{25.51\\(\textbf{10.0})}    & \makecell{40.03\\(10.0)}    & \makecell{49.19\\(10.0)} & \makecell{75.58\\(10.0)} & \makecell{11.27\\(1.01)} & \makecell{30.41\\(1.01)} & \makecell{43.51\\(1.01)} & \makecell{66.72\\(1.01)} & \makecell{55.83\\(15.57)} & \makecell{70.68\\(15.57)} & \makecell{74.53\\(15.57)} & \makecell{\textbf{83.38}\\(15.57)}\\
\hline
\makecell{Xavier\\Normal} &\makecell{26.37\\(8.45)} & \makecell{39.53\\(8.45)} & \makecell{48.67\\(8.45)} & \makecell{75.72\\(8.45)} & \makecell{11.28\\(0.93)} & \makecell{31.88\\(0.93)} & \makecell{43.81\\(0.93)} & \makecell{66.66\\(0.93)} & \makecell{55.17 \\(5.07)} & \makecell{70.76\\(5.07)} & \makecell{74.85\\(5.07)} & \makecell{83.10\\(5.07)} \\
\hline
Sylvester & \makecell{26.84\\(7.35)} & \makecell{\textbf{43.86}\\(\textbf{10.03})} & \makecell{\textbf{53.86}\\(\textbf{16.15})} & \makecell{\textbf{77.44}\\(\textbf{29.61})} & \makecell{\textbf{13.02}\\(\textbf{1.9})} & \makecell{\textbf{34.19}\\(\textbf{8.44})} & \makecell{\textbf{46.37}\\(\textbf{10.23})} & \makecell{\textbf{67.18}\\(\textbf{10.12})} & \makecell{\textbf{58.74} \\ (\textbf{41.27})} & \makecell{\textbf{72.90} \\ (\textbf{63.61})} & \makecell{\textbf{75.40} \\ (\textbf{65.33})} & \makecell{82.98 \\ (\textbf{66.06})} \\
\hline
\end{tabular}}
\vspace{-5pt}
\end{table}
From the table, we see that on the CIFAR-10 dataset, our proposed approach produces better test accuracy on all shot settings except the 10-shot case. This is mostly because in our Sylvester method for 10-shot case, we don't have enough data for producing better principal components. For the CIFAR-100 dataset, our proposed approach produces better test accuracy for all the shots. For the fine-tuning experiments, our method shows large improvement over the random methods especially for the lower shot settings but the gap reduces as the number of shots increases. This is because as the number of shots increase, there is no additional advantage of data-driven initialization as the random methods can reach better performance by gradient descent over larger number of samples. Also, the initial accuracy of our method is very close to the final accuracy of all random methods for all shots.
\subsection{Additional Analyses}
\begin{figure}[h]
    \vspace{-30pt}
    \centering
    \subfigure[]{\includegraphics[width=0.23\textwidth]{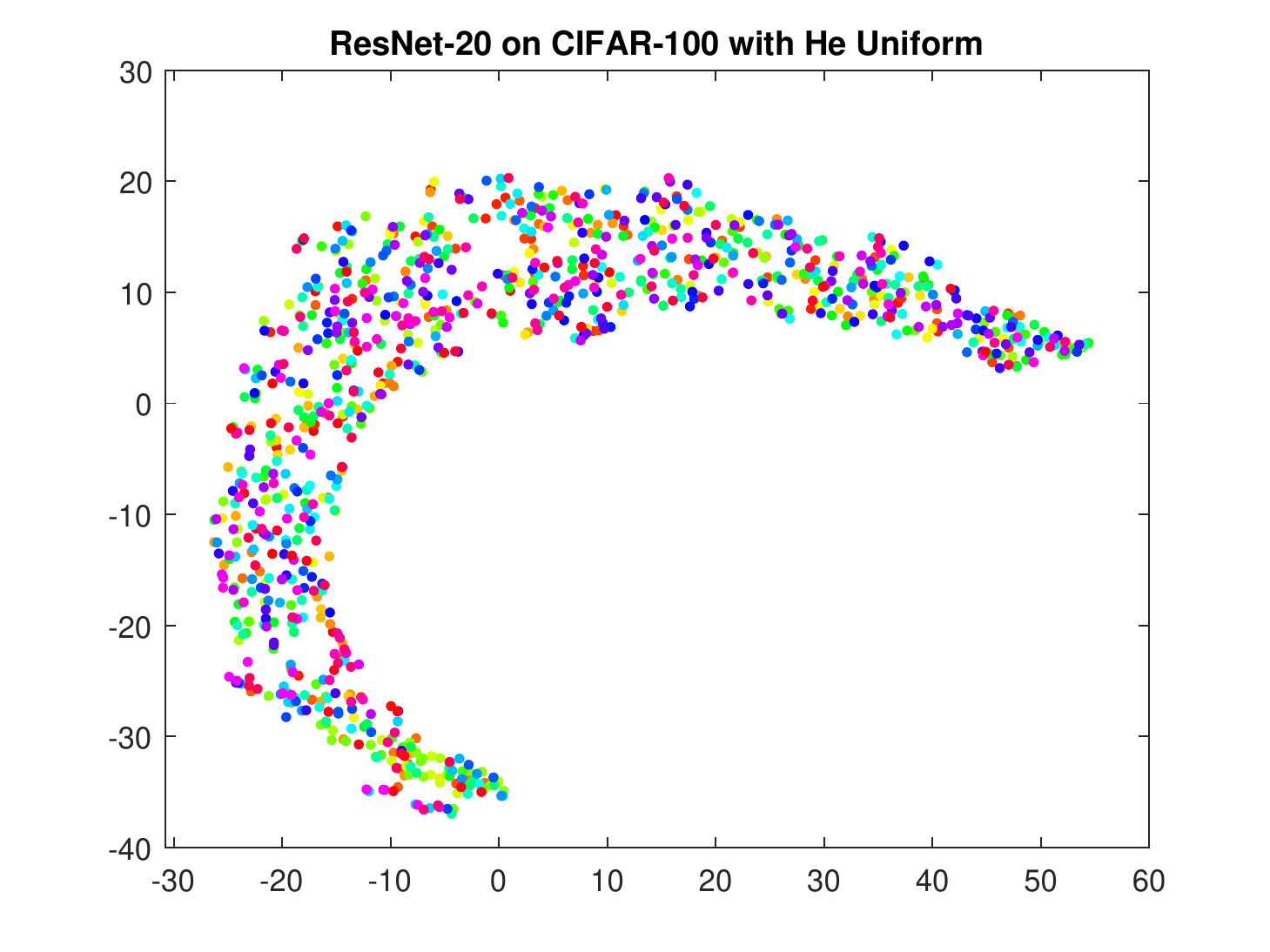}} 
    \subfigure[]{\includegraphics[width=0.23\textwidth]{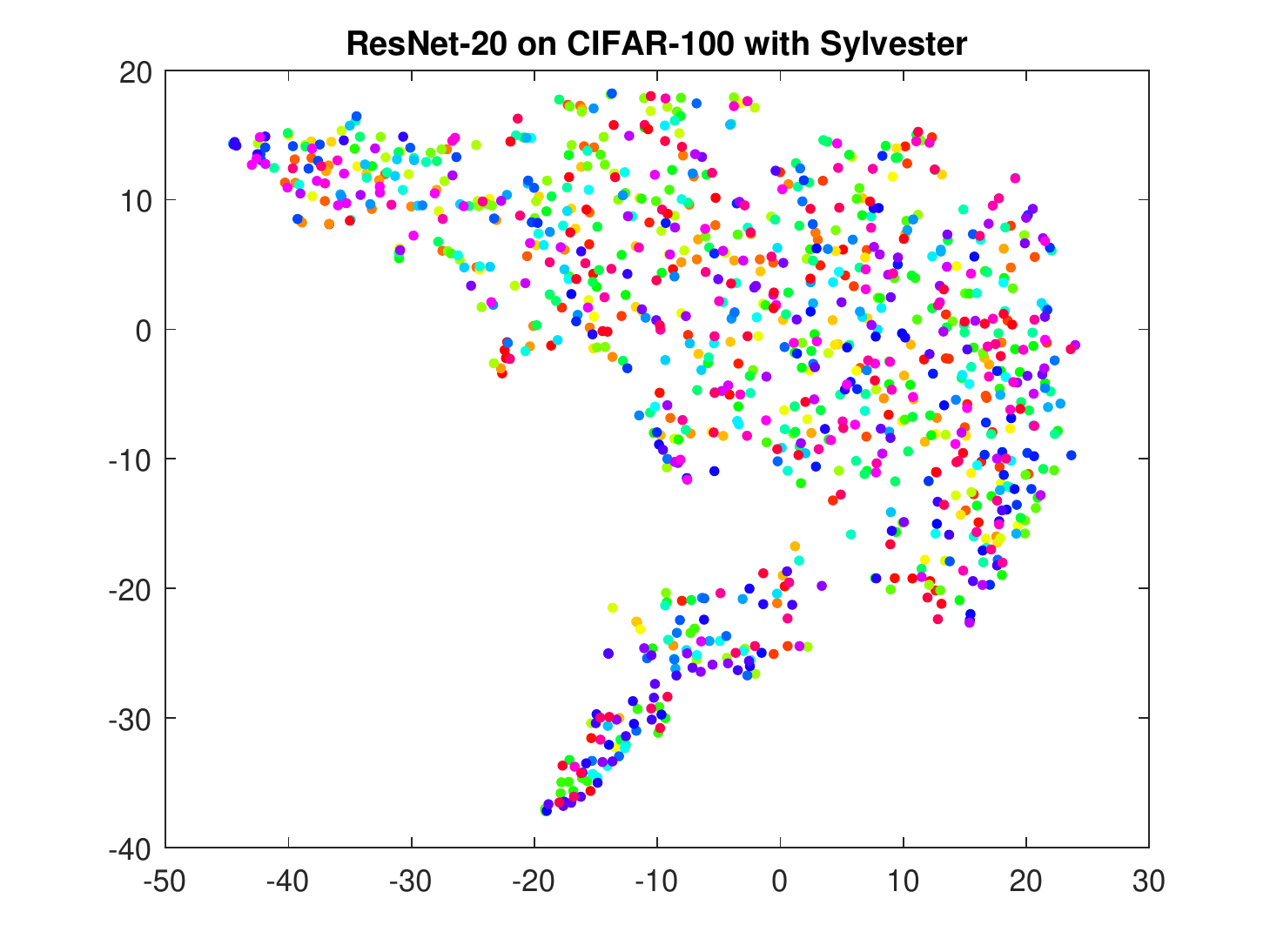}}
    \subfigure[]{\includegraphics[width=0.23\textwidth]{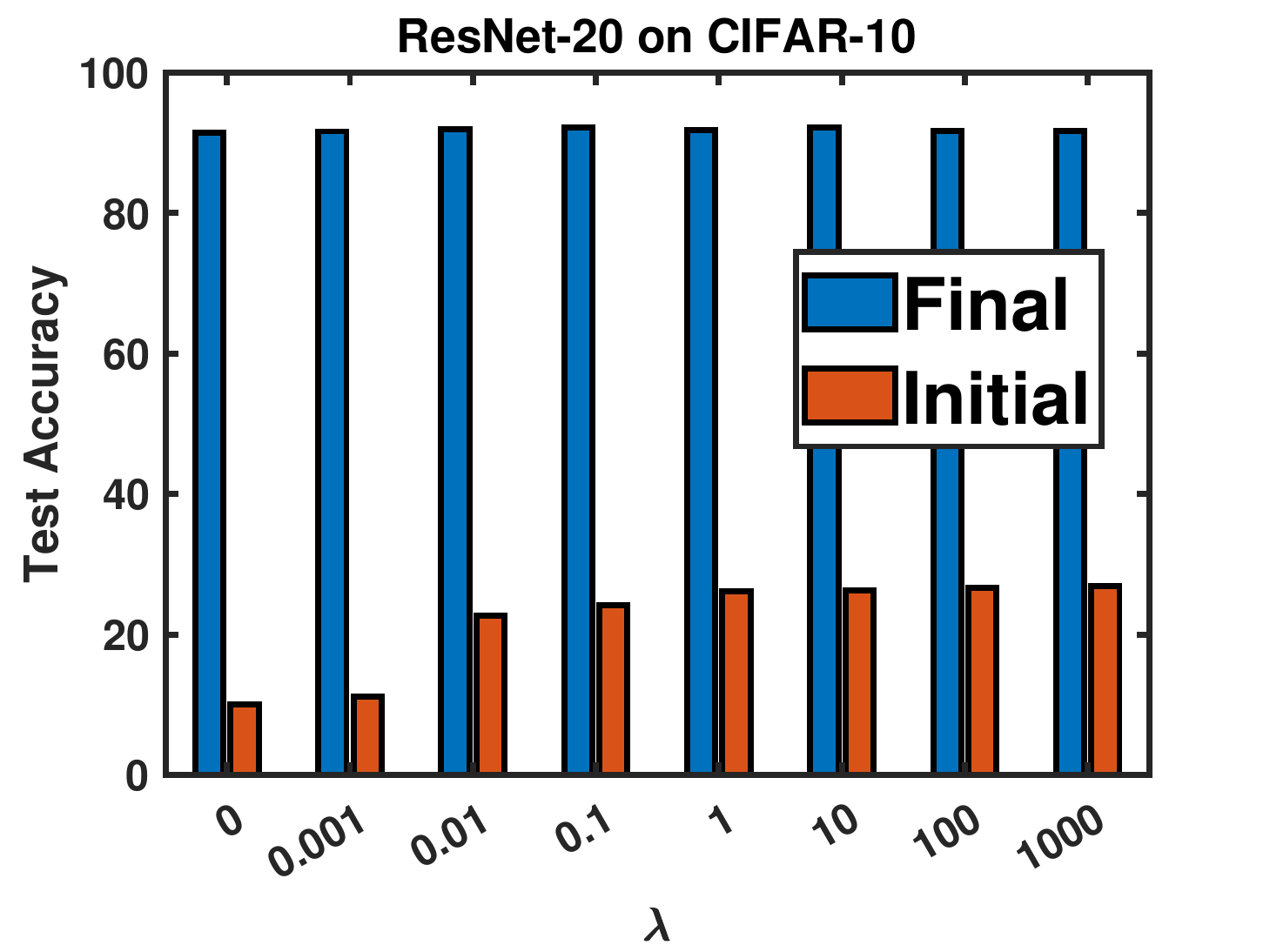}}
    \subfigure[]{\includegraphics[width=0.23\textwidth]{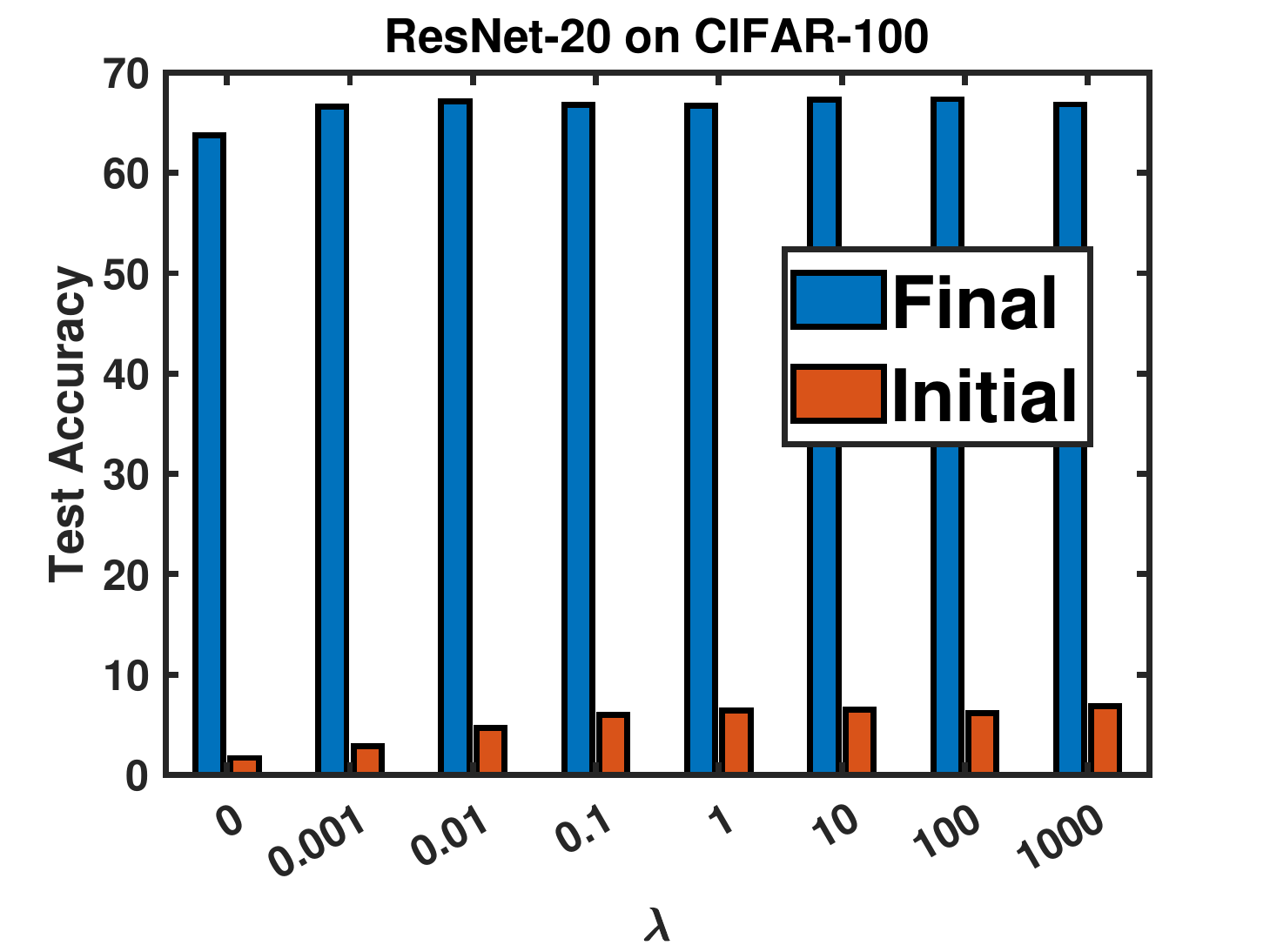}}
\vspace{-10pt}
    \caption{On the CIFAR-100 dataset, we obtain the following: (a) t-SNE plot of features when initialized with He Uniform; (b) t-SNE plot of features when initialized with Sylvester show cluster formation. Effect of $\lambda$ on initial and final accuracy for (c) CIFAR-10 and (d) CIFAR-100.}\label{fig:tsne}
    \vspace{0pt}
\end{figure}

We also further analyze our proposed method. We obtain 2D t-SNE~\citep{van2008visualizing} plots of the features after initialization and before training starts. Results for He uniform initialization on CIFAR-100 are shown in Fig.~\ref{fig:tsne} (a). Results for our Sylvester-based initialization on CIFAR-100 are shown in Fig.~\ref{fig:tsne} (b). The results show that there are no distinctive clusters when using the He uniform initialization. However, for our method, distinctive clusters begin to form for the CIFAR-100 dataset. Thus, it is visually justified why our method produces high initial accuracies compared to random methods.

Furthermore, we study the effect of $\lambda$ on test accuracy. $\lambda$ weighs the effect of encoding loss as described in~\eqref{eq:encdec}. The results on CIFAR-10 and CIFAR-100 datasets are shown in Fig.~\ref{fig:tsne} (c) and (d), respectively. The results show that increasing $\lambda$ generally increases the test accuracy. However, increasing $\lambda > 1$ produces saturation in performance. The plots also show that the contribution of the encoding loss is more compared to the decoding loss. This is mostly because eventually the encoder is used in the network and the decoder is just used for constraining the encoder to produce meaningful latent codes.
In Table~\ref{tab:sample}, we show the effect of number of samples on initialization time and recognition performance of our proposed approach for CIFAR-10 and CIFAR-100 datasets respectively. For CIFAR-10 dataset, the initial accuracy increases with number of samples but saturates at 300 samples per class. 100 seems to be the optimal number of samples required for initialization as recognition performance does not increase beyond that operating point. For CIFAR-100 dataset, the initial accuracy increases with increasing number of samples. The initialization time also increases with number of samples for both the dataset. Also, the initial accuracy is much better than that of He uniform for all the sampling settings on both the datasets.
  
\begin{figure}[!tbh]
    \vspace{-10pt}
    \centering
    \subfigure[]{\includegraphics[width=0.4\textwidth]{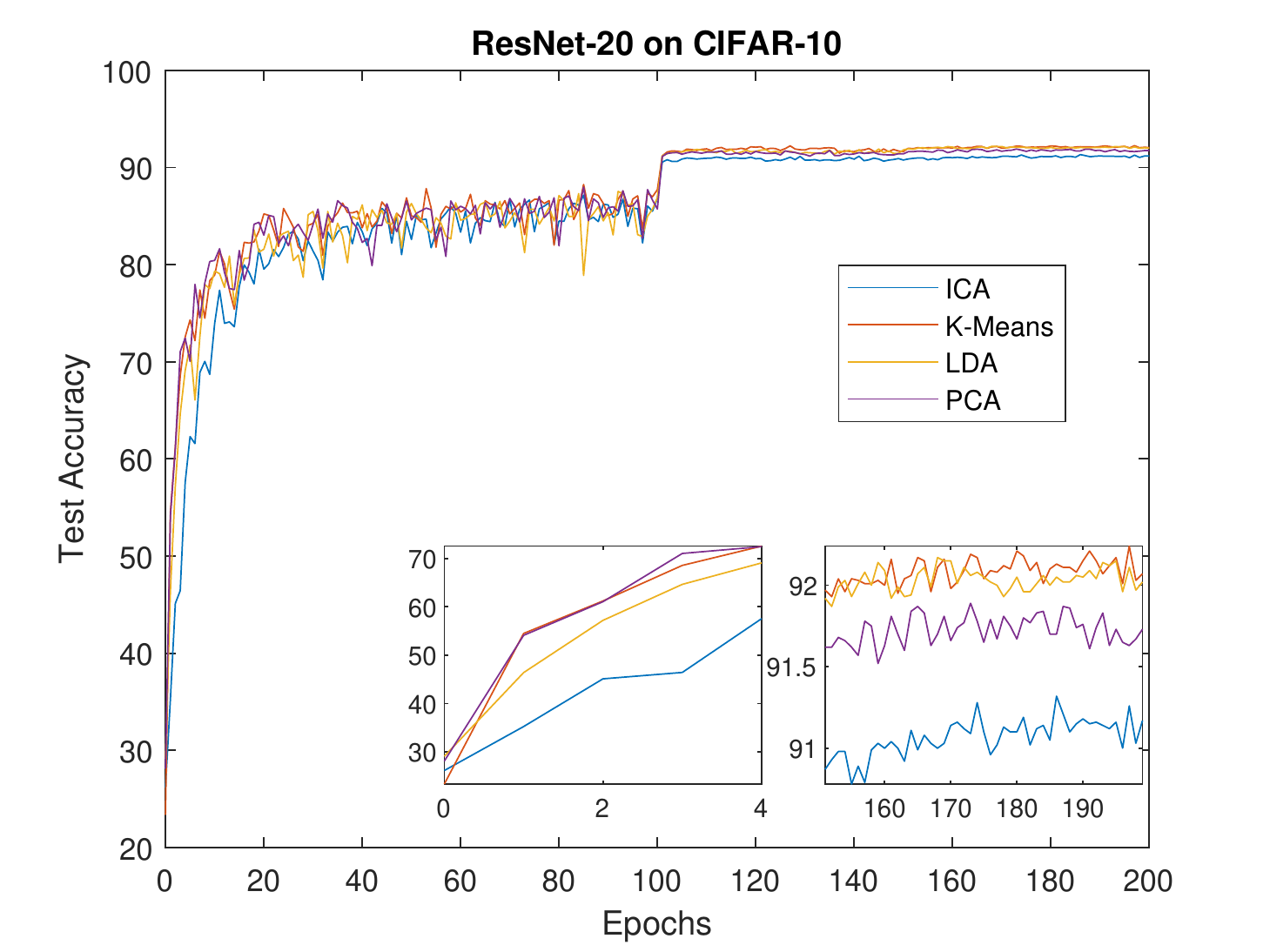}}
    \subfigure[]{\includegraphics[width=0.4\textwidth]{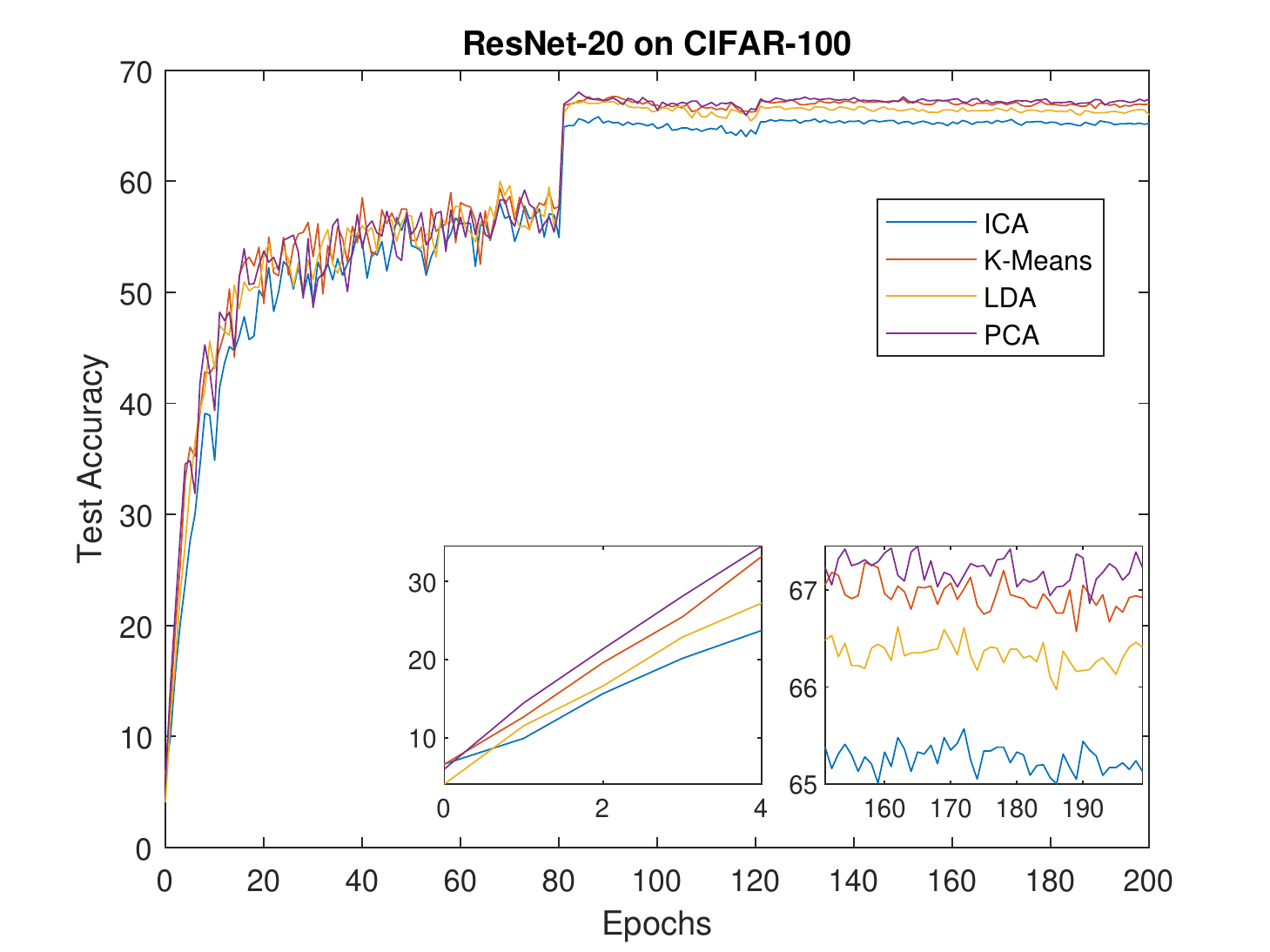}}\vspace{-10pt}
\caption{Effect of different target latent codes on (a) CIFAR-10 and (b) CIFAR-100 datasets.}\label{fig:targets}\vspace{-15pt}
\end{figure}

\begin{table}[h!]
\centering
\caption{Initialization time and initial accuracy of our proposed method for different sample counts per class as compared with He uniform (He-u) initializer}\label{tab:sample}
\vspace{0pt}
\resizebox{0.8\columnwidth}{!}{
\begin{tabular}{|c|c|c|c|c|c|c|c|c|}
\hline 
\textbf{Setup} &  \multicolumn{4}{c|}{\textbf{CIFAR-10}} &  \multicolumn{4}{c|}{\textbf{CIFAR-100}}\\
\hline
\makecell{Sample count per class $\rightarrow$} &  10 & 100 & 300 & He-u & 5 & 10 & 30 & He-u \\
\hline
\makecell{Time (s)} & \makecell{7.28} & \makecell{26.0} & \makecell{38.85} & \makecell{7e-4} & \makecell{43.7} & \makecell{82.2} & \makecell{242.07} & \makecell{7e-4}\\
\hline
\makecell{Accuracy (\%)} & \makecell{17.96} & \makecell{24.5} & \makecell{25.14} & \makecell{9.96} & \makecell{3.5} & \makecell{5.69} & \makecell{7.17} & \makecell{1.01} \\
\hline\end{tabular}}
\vspace{-5pt}
\end{table}
Finally, we study the effect of alternative latent codes ($\mathbf{S}$). We set the latent code $\mathbf{S}$ as the features obtained by applying Linear Discriminant Analysis (LDA) and Independent Component Analysis (ICA) on the input activations of each layer. We also use K-Means to obtain a latent code. For input activations $\mathbf{X} \in \mathbb{R}^{d_i \times n}$, we apply K-Means to obtain $d_o$ clusters arranged in the matrix $\mathbf{H} \in \mathbb{R}^{d_i \times d_o}$. Then, we apply inner product between the input activations and the cluster centers to obtain $\mathbf{S}=\mathbf{H}^T\mathbf{X}$. The $\mathbf{S}$ is then used to formulate the Sylvester equation and solve it to obtain the layer weights. The results of using these alternative latent codes are shown in Fig.~\ref{fig:targets} (a) and (b) on CIFAR-10 and CIFAR-100, respectively. The final test accuracy results show that for both the datasets, ICA seems to under-perform compared to other latent codes. This might be because the latent codes obtained using ICA do not represent features obtained using regular convolutional filters. On the other hand, the K-Means approach produces competitive final test accuracy with the best and the second best results on the CIFAR-10 and CIFAR-100 datasets respectively.

\section{Conclusion}
In this work, we proposed a data-driven initialization technique for feed-forward neural networks. Our method consists of a sequential approach where each layer is initialized and then the activations are propagated to facilitate initialization of next layer and so on. The weights of each layer are obtained by optimizing a combination of decoding and encoding loss which can be translated to solving the Sylvester equation. Experiments showed improved recognition performance of our approach compared to random methods especially in few-shot settings. The computation time of our method is higher compared to random methods and we plan to increase the efficiency of the solver. In the future, we would also like to test this initialization method on dense prediction tasks and on other datasets and architectures.

\bibliography{iclr2021_conference}
\bibliographystyle{iclr2021_conference}


\end{document}